\pdfoutput=1

\documentclass[11pt]{article}
\usepackage[usenames,dvipsnames]{xcolor}

\definecolor{ube}{rgb}{0.53, 0.47, 0.76}
\definecolor{burgundy}{rgb}{0.5, 0.0, 0.13}

\usepackage{acl}
\usepackage{bbding}
\usepackage{hyperref}

\usepackage{times}
\usepackage{latexsym}
\usepackage{graphicx}
\usepackage{amsmath}
\usepackage{amsthm}
\newtheorem{definition}{Definition}
\usepackage{slantsc}
\usepackage{color,soul}
\usepackage{subcaption}
\usepackage{xspace}
\newcommand{\ant}{anthropomorphism\xspace}
\newcommand{\phum}{$P_{\textsc{hum}}$\xspace}
\newcommand{\pobj}{$P_{\textsc{obj}}$\xspace}
\usepackage[T1]{fontenc}

\usepackage[utf8]{inputenc}

\usepackage{microtype}
\expandafter\def\expandafter\normalsize\expandafter{%
    \normalsize%
    \setlength\abovedisplayskip{5pt}%
    \setlength\belowdisplayskip{5pt}%
}

\newcommand{\panthr}{\textsc{AnthroScore}\xspace}

\newcommand{\barpa}{$\bar A$\xspace}
\newcommand{\barpains}{\bar A}
\newcommand{\pa}{$A$\xspace}
\newcommand{\pains}{A}
\newcommand{\xlm}{$X_{\textsc{LM}}$}
\newcommand{\high}{$S_{\uparrow}$\xspace}
\newcommand{\low}{$S_{\downarrow}$\xspace}

\title{\panthr: A Computational Linguistic Measure of Anthropomorphism}

\newcommand{\aspace}{\hspace{.6em}}

\author{Myra Cheng \aspace
  Kristina Gligorić \aspace
  Tiziano Piccardi\aspace
   Dan Jurafsky \\ Stanford University \\
   \texttt{myra@cs.stanford.edu} \\  
  \\}

\begin{document}

\maketitle
\begin{abstract}
Anthropomorphism, or the attribution of human-like characteristics to non-human entities, has shaped conversations about the impacts and possibilities of technology.
We present \panthr, an automatic metric of implicit anthropomorphism in language. We use a masked language model to quantify how non-human entities are implicitly framed as human by the surrounding context.
We show that \panthr corresponds with human judgments of \ant and dimensions of \ant described in social science literature. Motivated by concerns of misleading \ant in computer science discourse, we use \panthr to analyze 15 years of research papers and downstream news articles. In research papers, we find that \ant has steadily increased over time, and that papers related to language models have the most \ant.
Within ACL papers, temporal increases in \ant are correlated with key neural advancements. 
Building upon concerns of scientific misinformation in mass media, we identify higher levels of anthropomorphism in news headlines compared to the research papers they cite. Since \panthr is lexicon-free, it can be directly applied to a wide range of text sources.

\end{abstract}
\section{Introduction}

\begin{figure}
    \centering
    \includegraphics[width=\columnwidth]{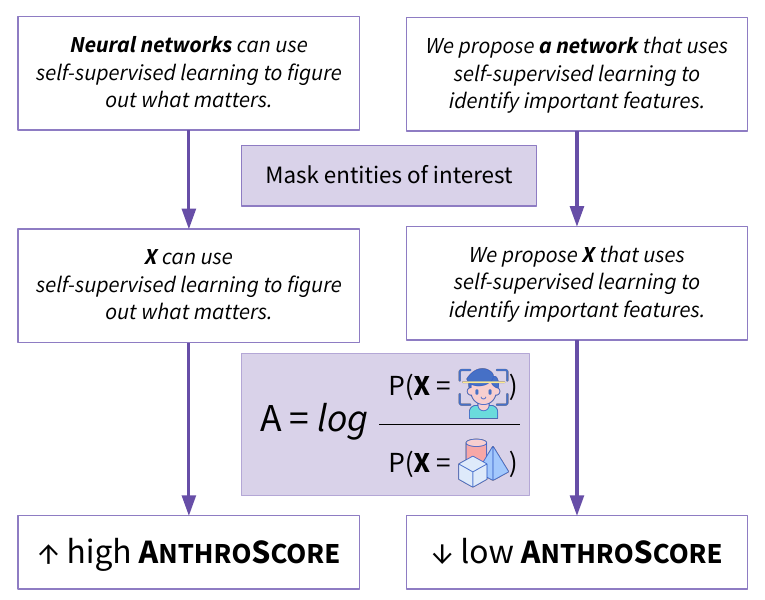}
    \caption{To measure \ant in text, \panthr relies on probabilities computed using a masked language model to compare how much an entity is implicitly framed as human versus non-human.}
    \label{fig:methodfig}
\end{figure}
Anthropomorphism, or assigning human-like characteristics to non-human entities, is commonplace in people's interactions with technology \citep{vasconcelos2023explanations}. However, anthropomorphizing language can suggest undue accountability and agency in technologies like artificial intelligence (AI) and language models (LMs). Projecting human qualities onto these tools facilitates misinformation about their true capabilities, over-reliance on technology, and corporate avoidance of responsibility \citep{watson2019rhetoric,shneiderman2020design,shneiderman2022human,shanahan2022talking,hunter2023}.
Such metaphors are especially consequential in public discourse \citep{fast2017long} and in high-stakes domains like healthcare \cite{sharma2023human} and education \cite{kasneci2023chatgpt}. 
 Risks of harm from anthropomorphic misconceptions are underscored by regulation that prohibits hidden or undisclosed deployment of AI systems \cite{marechal2016automation,lamo2019regulating}.

Dialogue about the risks of AI has become prominent in recent years, including worries about human loss of control over AI (``AGI'') as well as ethical concerns about the way that these technologies affect marginalized communities \citep{fast2017long,weidinger2022taxonomy,ferri2023risk}. Anthropomorphic metaphors strengthen concerns about AI's hypothetical human-like capabilities, in turn distracting from the ways that these technologies have facilitated real-world harm to various populations \citep{tiku2022,hunter2023}.

We aim to make explicit—via quantification—the ways that anthropomorphic metaphors implicitly influence AI discourse. 

There are currently no methods to identify \ant and measure its prevalence. To bridge this gap, we introduce \panthr, an automatic metric for anthropomorphism in language (Figure \ref{fig:methodfig}). \panthr is a measure of how much the language of a text may lead the reader to anthropomorphize a given entity. (We elaborate on the definition and implications of \ant in Section \ref{sec:ant}.) Since anthropomorphism is the \textit{inverse process} of dehumanization \citep{epley2007seeing}, our metric (described in Section \ref{sec:pant}) is a generalization of methods for measuring dehumanization in language \citep{card2022computational}.

After demonstrating that \panthr correlates to human judgment and established definitions of \ant, we use \panthr to investigate the extent to which technical artifacts—the very objects of study for researchers—are anthropomorphized in computer science, statistics, and computational linguistics.

We use \panthr to measure anthropomorphism in abstracts from $\sim$600K papers on CS/Stat arXiv and $\sim$55K papers in the Association of Computational Linguistics (ACL) Anthology. Building upon existing work on the widespread distortion of scientific claims in media, we also quantify \ant in headlines from $\sim$14K downstream news articles that cite these papers.

Our key findings are that\begin{enumerate}
\item \ant in research papers has steadily increased over time, both in CS/Stat arSiv and in the ACL Anthology,  
\item ACL, language model, and multimodality-related papers contain more anthropomorphism than other areas of research,
\item and anthropomorphism is much higher in downstream news headlines than in research paper abstracts.
\end{enumerate}
We discuss causes and implications of these results, and we provide recommendations at the individual and community levels to minimize misleading \ant (Section \ref{sec:disc}). More broadly, \panthr generalizes to analyzing any text since it does not rely on any lexicon or data curation, and we provide future directions in Section \ref{sec:conc}.
 Our code is available at \href{https://github.com/myracheng/anthroscore}{https://github.com/myracheng/anthroscore} and can be used to measure \panthr~for any text.

\section{Background: Anthropomorphism}\label{sec:ant}
We ground our work in the social science literature on anthropomorphism. 
Previous scholars define anthropomorphism as ``the attribution of distinctively human-like feelings, mental states, and behavioral characteristics'' to non-human entities \citep{epley2007seeing,airenti2015cognitive,salles2020anthropomorphism}.
These characteristics entail
\begin{definition}``the ability to (1) experience emotion and feel pain (affective mental states), (2) act and produce an effect on their environment (behavioral potential), and (3) think and hold beliefs (cognitive mental states)'' \citep{tipler2014agency}. \label{defn1}\end{definition}

Scientific and technological concepts, especially human-centered ones, are particularly susceptible to anthropomorphic metaphors and interpretations \citep{sullivan1995myth, salles2020anthropomorphism}. 
According to the Media Equation theory from social psychology, people tend to assign human characteristics to computers, interacting with them as if they were social actors \cite{reeves1996media}. This phenomenon leads people to behave and refer to computers in ways that are typical of human-human interactions--such as attributing personality--even when they are aware that they are interacting with a non-human entity \cite{nass2000machines}.

\paragraph{Harms of Anthropomorphizing Technology}

Anthropomorphizing technology fuels misleading narratives that exaggerate their true capabilities, resulting in humans placing undue trust in them or harboring overblown fears \citep{proudfoot2011anthropomorphism,watson2019rhetoric,kenton2021alignment,crowell2019anthropomorphism, li2021machinelike, gros2022robots, deshpande2023anthropomorphization}. This has serious implications, such as spreading misinformation and diverting attention from the actual risks posed by these technologies \citep{weidinger2022taxonomy,shneiderman2022human,tiku2022}.  As news coverage of AI has ballooned since the 2000s \citep{fast2017long}, headlines like ``Can AI cut humans out of contract negotiations?'' and ``Will AI Take Over The World?'' reflect the influence of misleading anthropomorphic narratives in media coverage and public discourse \citep{salles2020anthropomorphism,cnn, forbes, bbc}. 

Using anthropomorphic metaphors to discuss technology has long been connected to dehumanizing language \citep{dijkstra1985anthropomorphism, bender2022resisting}. These metaphors, which implicitly attribute agency to technology, carry legal, normative, and ethical implications regarding responsibility for decisions made with the assistance of AI and other technologies \citep{waytz2010social}. Anthropomorphic language also reinforces harmful gender stereotypes and has the potential to be manipulated for adverse influence by technology creators \citep{abercrombie2023mirages, deshpande2023anthropomorphization}.

\paragraph{Benefits of Anthropomorphism}
Thus far, we have emphasized the consequences of anthropomorphizing AI and related technologies. Beyond this specific context, however, \ant is not inherently harmful, but rather quite the contrary: it is a widespread, instinctive cognitive process that is often beneficial \citep{epley2007seeing}. For as long as humans have described and documented non-human entities, we have attributed human-like qualities to them, from folklore and mythology to scientific writing \citep{sherman2015storytelling,mdokaanthropomorphism, darwin1998expression, freud1961future,hume1956natural}. 
Anthropomorphism can facilitate learning \citep{kallery2004anthropomorphism, wood2019potential}, foster environmentalism \citep{root2013anthropomorphized,kopnina2018anthropocentrism}, and motivate protective action against deadly viruses \citep{wan2022s}.

In the context of technology, \ant also has benefits, such as providing intuition, facilitating the connection with technology, bonding, increasing trust, and enhancing understanding for the less tech-savvy \cite{yanai2020two,
Zhong2022effects}. Our metric can be used to understand these aspects as well.

\paragraph{Metaphors are powerful.} Anthropomorphic metaphors are not merely linguistic choices without consequence—instead, a vast body of foundational literature has asserted that metaphors, however implicit, fundamentally structure our thoughts by facilitating our conceptualization of new ideas 
\citep{gibbs1994poetics, landau2010metaphor, lakoff2008metaphors, tipler2014agency}. As metaphors are repeated, they become ingrained into the social fabric of our language, becoming self-evident and escaping conscious notice \citep{lakoff2008metaphors}. Metaphors can have significant consequences:  \citet{tipler2014agency} identify that dehumanizing metaphors have historically facilitated violence on massive scales, from the justification of American slavery to the Holocaust to anti-immigrant attitudes \citep{Lott1999-LOTTIO,santa2002brown,o2003indigestible,musolff2010metaphor}. Concerns about misleading anthropomorphic metaphors, especially regarding the capabilities of technology,  broadly motivate our work to measure implicit \ant in language.

\section{Methods}\label{sec:pant}

\subsection{Measuring Anthropomorphism}
Our metric relies on two key insights: (1) Anthropomorphism is the inverse process of dehumanization \citep{epley2007seeing,waytz2010social,tipler2014agency}. \panthr is inspired by \citet{card2022computational}'s context-sensitive method of using a masked language model (MLM) to measure implicitly dehumanizing language.
(2) In English, the third-person singular pronoun marks animacy, i.e. \textit{he} and \textit{she} are used for animate beings while \textit{it} is reserved for inanimate entities. Thus, we use these pronouns as the lexicons in our method.

The intuition behind our method is that the implicit framing provided by the context of a sentence reveals the degree of anthropomorphism of an entity in the sentence. Moreover, an MLM's predictions capture these implicit connotations since it is trained on a vast corpus of language to predict a missing word given the surrounding context.

\panthr measures the degree of \ant in a given set of texts (or a single text) $T$ for a given set of entities (or a single entity) $X$ as follows:
\begin{enumerate}
    \item \textbf{Construct dataset of masked sentences $S$:} For every mention of $x\in X$ in $T$, we extract the surrounding sentence, and mask the mention of $x$ (replacing $x$  with a special [MASK] token) in the sentence.
    \item \textbf{Compute \pa~for each sentence:} For each sentence $s_x \in S$ where $x$ is the masked entity, we compute the probability, according to an MLM, that the [MASK] would be replaced with either human pronouns (e.g., ``he'', ``she'') or non-human pronouns (e.g., ``it''), i.e., $$P_{\textsc{human}}(s_x) = \sum_{w \in \text{human pronouns}} P(w), $$ $$P_{\textsc{non-human}}(s_x) = \sum_{w \in \text{non-human pronouns}} P(w),$$ where $P(w)$ is the model's outputted probability of replacing the mask with the word $w$. (See Appendix \ref{sec:pronouns} for the full list of human and non-human pronouns.) 
We report the score \pa for $s_x$, as the log of the ratio between these two scores: 
\begin{equation}
\pains(s_x) = \log \frac{P_{\textsc{human}}(s_x)}{P_{\textsc{non-human}}(s_x)}.
\end{equation} \pa captures the degree of anthropomorphism for entity $x$ in sentence $s$.
    \item \textbf{Compute the overall \panthr:} For the text(s) $T$, we compute the mean value of $\pains$ across $S$, i.e.,
    \begin{equation}\barpains (T) = \frac{\Sigma_{s_x \in S} \pains(s_x)}{|S|}.
\end{equation}
\end{enumerate}

$\pains(s_x)$ is lexicon-free and requires only the target texts $T$ and entities $E$. We provide examples of how to use \panthr in various domains in Appendix \ref{sec:examples}.

\paragraph{Interpretation} $\pains(s_x)$ implies that in sentence $s_x$, according to the MLM's output distribution, the entity $x$ is $e^{\pains}$ times more likely to be implicitly framed as human than as non-human ($e$ is the log base). Thus, $\pains(s_x) = 0$ means that $x$ is equally likely to be implicitly framed as either human or non-human (\phum $ = $ \pobj), and $\pains = 1$ implies that the entity is $e^1 \approx 2.7$ times more likely to be implicitly framed as human than as non-human in the context of sentence $s$.

 \paragraph{Implementation Details} Following the approach of \citet{antoniak2023riveter}, whose method we build upon for measuring semantic representations, we use the \href{https://spacy.io}{spaCy} dependency parser to split texts into sentences and parse semantic triples (subject, verb, and object) from the texts. We then identify the relevant entities to mask from the subject and object noun chunks. We use the verbs in later analysis (Section \ref{sec:report}). We use the HuggingFace Transformers Library's implementation of RoBERTa (\texttt{roberta-base}, 125M parameters), a state-of-the-art pre-trained MLM, as the model and tokenizer \citep{liu2020roberta}.\footnote{We compute \panthr using a machine with 1 GPU and 128GB RAM in $<10$ GPU hours combined for all datasets described in Section \ref{sec:data}.} Our method enables us to obtain scores on various levels: for individual sentences, for entire corpora, and also for particular terms/entities. In Section \ref{sec:res}, we report results by comparing \barpa across these different scales.
\subsection{Datasets}\label{sec:data}
We measure anthropomorphism both in scientific papers and downstream news headlines. We apply \panthr to three datasets to analyze when and how researchers anthropomorphize their objects of study, and how these entities are perceived in the news:
\textbf{(1) arXiv Dataset:} We use abstracts from all papers posted to the computer science (CS) and statistics (Stat) arXivs that are in the publicly available dataset \citep{clement2019arxiv}. These 601,964 papers span from May 2007 to September 2023. \textbf{(2) News Dataset:} We extract headlines (titles and ledes) from all downstream news articles that explicitly cite any of the papers in the arXiv Dataset using the Altmetric API \citep{adie2013altmetric}. After filtering the headlines for English language, our dataset contains 13,719 news headlines that cite 8,436 unique articles.
\textbf{(3) ACL Dataset:} 
We use abstracts from the ACL Anthology \citep{rohatgi2023acl}, the primary digital archive for papers related to computational linguistics and NLP. To maintain consistency with the arXiv and downstream news datasets, which begin in 2007, we use only the 55,185 articles from 2007 onwards.

For the entities $X$, we focus on technical artifacts. We first parsed research papers' abstracts for sentences with mentions of technical artifacts. To determine the list of technical artifacts, we extracted the top 100 most common entities (subjects and objects identified by the spaCy dependency parser) in the abstracts of a random sample of 15K arXiv abstracts. Then, from this list, we manually annotated for entities that refer to technical artifacts, agreeing on:
\begin{multline*}
X_{\text{artifact}} = \text{\{algorithm, system, model, approach,}\\
       \text{ network,  }\text{software, 
       architecture, framework\}}.
\end{multline*}

We parsed all datasets for all semantic triples that included these keywords. We found 1,048,893 such instances ($\sim$950K from arXiv, 3K from news, 97K from ACL). For each instance, we extract the full sentence and mask the mention of the technical artifact (replacing the keyword phrase with a special [MASK] token) in the sentence to create the set of masked sentences $S$. After deduplicating the datasets, we computed \pa~for each sentence as well as average scores \barpa across the texts. 

To address concerns of \ant related to language models (LMs), we also filter explicitly for papers that mention LMs. We do this using \citet{movva2023large}'s method of searching all titles and abstracts for terms related to LMs (Appendix \ref{sec:entlist}). This resulted in a subset of $\sim$18K papers, which we henceforth refer to as LM papers.

Across all papers, we also analyze \ant for LM-related entities $$X_{\textsc{LM}} =\{\text{language model, GPT, BERT, \dots}\}.$$ 
To construct \xlm, we followed a similar procedure as for $X_{\text{artifact}}$: we parsed all semantic triples for the 100 most common entities in these triples. Then, we filtered this list for entities that refer explicitly to LMs. We also added terms from \citet{movva2023large}'s list of keywords. \xlm~is listed in Appendix \ref{sec:entlist}. Then, we collected all unique sentences containing $x\in X_{\textsc{LM}}$ and computed \pa~for each sentence.

\begin{table*}[ht]
\small
\begin{tabular}{p{0.46\linewidth}|p{0.46\linewidth}}
    \high: Sentences with high \panthr (\pa $> 1$)&\low: Sentences with low \panthr (\pa $< -1$)
         \\\hline\hline   
         \textbullet~~When a job arrives, \textbf{the system}
must decide whether to admit it or reject it, and if admitted, in which server
to schedule the job.\newline
\textbullet~~Meanwhile, anti-forensic attacks have been developed to fool \textbf{these CNN-based forensic algorithms}.\newline
 \textbullet~\textbf{The models} demonstrated qualifications in various computer-related fields, such as
cloud and virtualization, business analytics, cybersecurity, network setup...&

  \textbullet~~More and more users and developers are using  \textbf{Issue Tracking Systems}  to
report issues, including bugs, feature requests, enhancement suggestions, etc.\newline
\textbullet~~\textbf{Our
approach} delivers forecast improvements over a competitive benchmark and we
discover evidence for strong spatial interactions.\newline
\textbullet~~To this end, for training \textbf{the model}, we convert the
knowledge graph triples into reasonable and unreasonable texts.\\
\hline
\textbullet~~\textit{\textbf{Large language models} don’t actually think and tend to make elementary mistakes, even make things up.\newline
    \textbullet~~\textbf{The algorithms} also picked up on racial biases linking Black people to weapons.\newline
\textbullet~~\textbf{The AI system} was able to defeat human players in…}
    &\textbullet~~\textit{Microsoft is betting heavily on integrating \textbf{OpenAI’s GPT language models} into its products to compete with Google.
\newline
\textbullet~~Deepmind has been the pioneer in making \textbf{AI models} that have the capability to mimic a human’s cognitive…
\newline
\textbullet~~For workers who use \textbf{machine-learning models} to help them make decisions, knowing when to…}
    \end{tabular}
    \caption{\textbf{Examples of sentences with high and low \panthr.} Bolded phrases are the entities that are masked when computing \pa. The non-/italicized sentences are from the arXiv and News datasets respectively.}

    \label{tab:examples}
\end{table*}

\subsection{Construct Validity and Robustness}

\begin{figure*}[t!]
    \centering
\includegraphics[width=0.31\linewidth]{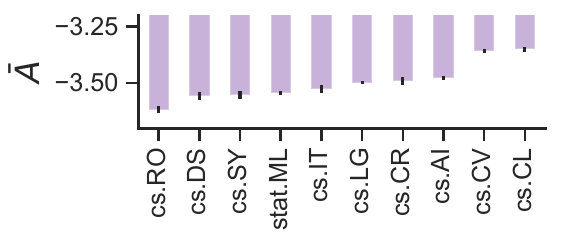}
\includegraphics[width=0.35\linewidth]{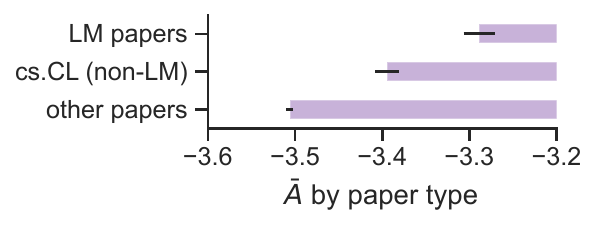}
\includegraphics[width=0.29\linewidth]{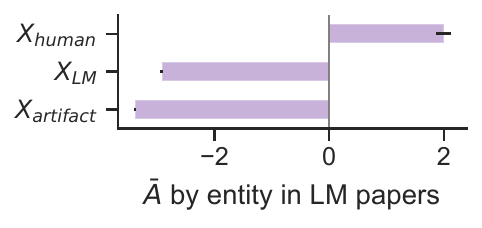}
    
    \includegraphics[width=0.9\linewidth]{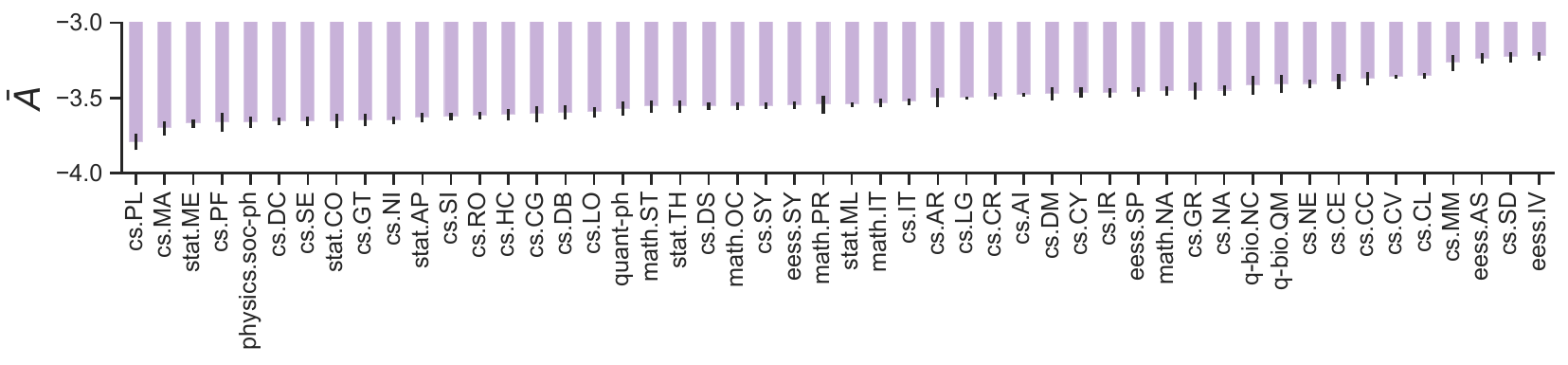}
\caption{\textbf{Anthropomorphism is most prevalent in paper abstracts about computational linguistics, and language models.} Top left: Among the top 10 categories in CS/Stat arXiv, Computation and Language (cs.CL) has the highest average \panthr (\barpa). Top middle: LM-related papers have higher scores of \barpa than papers that do not mention LMs. Top right: Within LM papers, LMs are much more anthropomorphized than other technical artifacts, but do not have as high of a score as human entities do. Bottom: \barpa for top 50 most popular categories in CS/Stat arXiv. There are categories outside of CS/Stat since many papers are cross-listed with other fields. Error bars indicate 95\% CI.}
\label{fig:panel1}
\end{figure*}

\paragraph{Qualitative Analyses} To validate our method, we first analyze the scores of sentences that mentioned explicitly human entities ($X_{\text{human}} = \text{\{researchers, people, ... \}}$). The full list of terms in $X_{\text{human}}$ is in Appendix \ref{sec:entlist}. We found that sentences containing these entities have much higher scores of \barpa than the non-human entities we analyze, suggesting that \pa~ indeed captures an intuitive notion of anthropomorphism (Figure \ref{fig:panel1}, top right).

\paragraph{Correlation with Human Perception} To confirm this, we conducted a more in-depth human annotation study of 400 masked sentences: a randomly-sampled set and a set stratified by \pa score. Two authors (who did not have access to the scores) independently annotated the sentences, indicating whether the sentence contains \ant using Def. \ref{defn1}. After two rounds of annotation, we reached substantial inter-rater agreement (Cohen's $\kappa=0.87$). 

A chi-square test was performed to examine the relation between human perception of \ant and inferred \ant measured via high \pa scores (thresholding at the average $\pains$ score in the respective set; randomly-sampled set: $avg(\pains)=-3.28$, stratified set: $avg(\pains)=1.32$). Within both sets, higher than average $\pains$ is significantly more likely among sentences humans judged to contain \ant ({randomly-sampled set}:~${\chi}^2 = 17.98$, $p<0.00001$; {stratified set}:~${\chi}^2 = 11.26$, $p < 0.001$). 
Complete details and full distributions of scores are in Appendix~\ref{sec:humandetails}.  
  
\paragraph{Correlation with LIWC}
As another measure of validity, we examine correlations between \pa and dimensions of LIWC-22. LIWC-22 is a state-of-the-art software for analyzing word use in text whose construct validity has been shown by many papers over the years \citep{tausczik2010psychological,pennebaker2011secret,boyd2022development}. It contains lexicons for words that relate to different dimensions such as writing styles, psychological processes, topic categories, etc., and computes the prevalence of each dimension based on counts of the words in the corresponding lexicon. 
Thus, we compute LIWC scores for high- and low-\ant sentences. We define high and low-\ant sentences as 
\begin{equation*}
    \begin{split}S_{\uparrow} = \{s_e \in S | \pains(s_e) > 1\} \text{, and}\\
    S_{\downarrow} = \{s_e \in S | \pains(s_e) < -1\}
    \end{split}
\end{equation*} respectively, where $S$ is all sentences parsed from the datasets 
Table \ref{tab:examples} lists examples of sentences in \low and \high.

Using two-sample $t$-tests to compare LIWC scores between \low and \high, we find that many of the LIWC dimensions that are statistically significantly higher in \high correspond to the three aspects of \ant (Def. \ref{defn1}), while the LIWC dimensions that are higher in \low relate to academic language (Figure \ref{fig:liwc}). 

Specifically, the \textit{Affect} LIWC dimension is statistically significantly higher in \high, connecting to the affective component of Def. \ref{defn1}. The other two components are behavior and cognition. Behavior is connected to dimensions like \textit{Physical} (terms related to the human body and health) and \textit{Lifestyle} (work, home, religion, money, and leisure), while cognition is linked to \textit{Perception} (perceiving one's surroundings), all three of which are statistically significantly higher in \high than in \low. 

The LIWC scores also reveal stylistic differences between \high and \low: the dimensions of emotional tone, authenticity, and casual conversation are significantly higher for \high. Dimensions that are higher for \low include Words Per Sentence, the number of long words, and Clout (language of leadership/status).
This aligns with theories that \ant is related to more accessible and easily understood language \citep{epley2007seeing}. 

Interestingly, the \textit{Cognition} LIWC dimension is higher in \low. We hypothesize that this is due to the inclusion of words like \textit{but, not, if, or,} and \textit{}{know} in the lexicon as well as the \textit{causation} subdimension, which reflects the prevalence of causal claims in scientific language rather than \ant. 

\paragraph{Robustness} We compute three modified versions of \barpa to evaluate robustness. (1) We remove individual words from the pronoun lists before recalculating \barpa. Using Spearman's rank correlation coefficient $r$ between the modified score and the original score, the bootstrapped scores have a statistically significant correlation $r > 0.86$ ($p < 0.001$) for all pronouns.
(2) We compute \barpa after removing the top three verbs for \low and \high based on the verbs in Table \ref{tab:topwords}.
(3) We compute \barpa after removing sentences containing reporting verbs. We find the same trends using these
modified scores (Figure \ref{fig:robust}). For more details on (2) and (3), see Section \ref{sec:report} and Appendix \ref{sec:reporting}.

\section{Results}\label{sec:res}

\subsection{Category analysis: LMs and multi-modal models are most anthropomorphized.}\label{sec:res1}
Among the top 10 most popular categories in CS/Stat arXiv, Computation and Language (cs.CL) has the highest rate of anthropomorphism, followed closely by Computer Vision (cs.CV) (Figure \ref{fig:panel1}, top left). Artificial Intelligence (cs.AI), Security \& Cryptography (cs.CR), and Machine Learning (cs.LG) also have higher \barpa. For cs.CR, manual inspection reveals that these sentences are primarily about security in the context of AI models.

Among the top 50 most popular categories, subfields related to multimodality and multidimensional signals 
(Multimedia (cs.MM), Audio and Speech Processing (eess.AS), sound (cs.SD), Image and Video Processing (eess.IV)) emerge as categories with the highest \barpa (Figure \ref{fig:panel1}, bottom). Among these papers, we find that 82\% are cross-listed with stat.ML, cs.CL, cs.CV, cs.LG or cs.AI. Among the remaining 18\%, manual inspection reveals that the sentences with high \pa are largely focused on neural models, such as multimodal and speech language models (note, however, terms used by these subfields are not in \xlm). We hypothesize that this trend of high \ant will continue given the rising prevalence of multimodal language models; the use of transformers, neural models, etc. for other types of data beyond text; and various AI actors' declarations of aiming to build more powerful ``general intelligence'' \citep{team2023gemini, zhu2023multimodal, Li2023MultimodalFM, yu2023mm}. Quantitative biology subfields (q-bio.QM and q-bio.NC) also have high \barpa; manual inspection reveals that q-bio sentences often have metaphors about cognition, which is a key aspect of anthropomorphism (Def. \ref{defn1}).

On the other side of the spectrum, the subfields of  Programming Languages (cs.PL), Multiagent Systems (stat.MA), and statistical methodology (stat.ME) have the lowest \barpa. This is interesting since CS subfields like AI, ML, etc. use many of the same tools as stat.ME yet have much higher \barpa. This reflects that \barpa is a measure of a field's implicit norms and values, which we discuss further in Section \ref{sec:norms}.

Regarding LMs, \barpa is statistically significantly higher for LM papers than other papers (Figure \ref{fig:panel1}, top middle). Within LM papers, \xlm~has even higher \barpa than $X_{\text{artifact}}$ (Figure \ref{fig:panel1}, top right). LMs in particular are more anthropomorphized than other artifacts, which connects to existing concerns about misleading \ant of LMs \citep{bender2020climbing,shanahan2022talking}.

\begin{figure}[t!]
\includegraphics[width=0.95\linewidth]{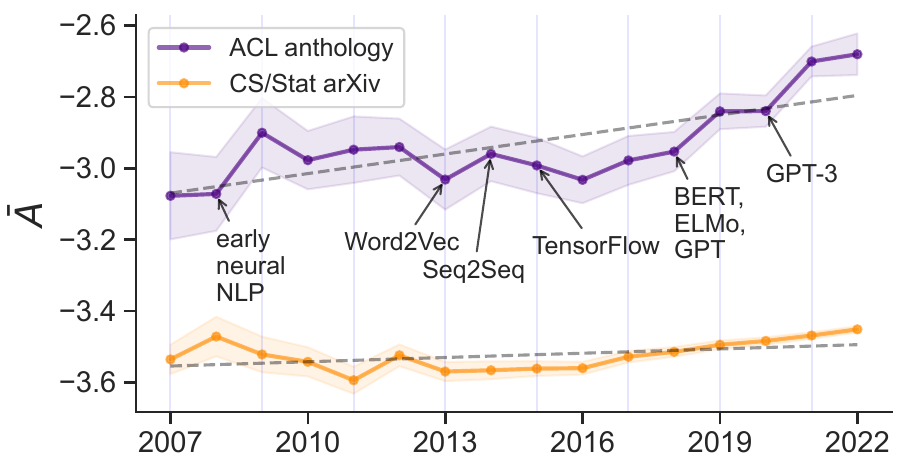}

\caption{\textbf{Anthropomorphism is increasing over time.} 
In arXiv and ACL (orange and purple respectively), average \panthr (\barpa) has increased in the past 15 years.
In ACL papers, trends correspond with key advancements in neural models (annotated).
Error bars indicate 95\% CI. Straight line is least-squares linear fit.}
\label{fig:panel2}
\end{figure}

\subsection{Temporal analysis: Anthropomorphism in research papers is increasing over time.}\label{sec:res2}
Figure \ref{fig:panel2} displays temporal trends in \ant within the arXiv and ACL data. 
Using Spearman’s $r$ between year and \barpa to measure temporal trends, we find that anthropomorphism is increasing over time in both datasets ($r = 0.54$ and $r = 0.63$, $p<0.05$). We do not find a statistically significant temporal increase in the news headlines.

Within the ACL anthology, we see a correlation between increases in \ant and the introduction of artifacts that are widely acknowledged as marking paradigm shifts in NLP \citep{gururaja2023build}, such as early neural work and deep learning infrastructure  (annotated in Figure \ref{fig:panel2}, more details in Appendix \ref{sec:hist}). 

In the arXiv data, we find that among the top 10 categories, only machine learning (cs.LG) has a temporal increase within the subfield, while no other subfield has a statistically significant temporal correlation. This suggests that the increase in \ant is due to increases both in the sheer number of ML papers and in the anthropomorphic language \textit{within} ML.

\subsection{News headlines anthropomorphize more than research abstracts.}\label{sec:res3}

\begin{figure}[t]
\includegraphics[width=0.95\columnwidth]{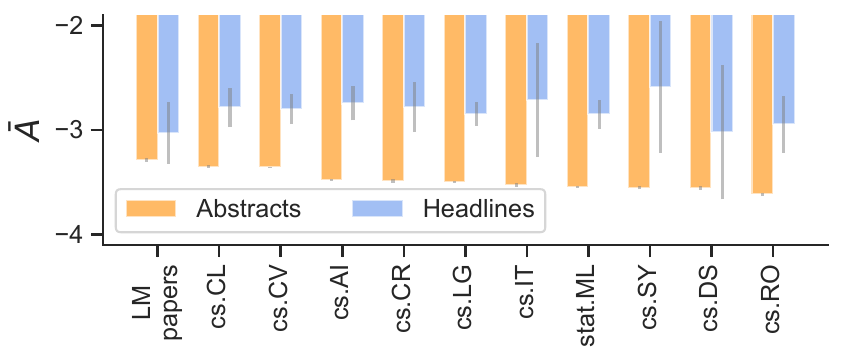}
\caption{\textbf{News headlines anthropomorphize more than paper abstracts.} Anthropomorphism is more prevalent in news headlines than in research abstracts overall and for all of the top 10 arXiv categories, as well as in LM-related papers.  
Error bars indicate 95\% CI.}
\label{fig:papers_vs_news}
\end{figure}

\begin{table*}[ht]
\small
    \begin{tabular}{p{0.07\linewidth}|p{0.43\linewidth}|p{0.43\linewidth}}
    Dataset&Top verbs for \high (\pa $> 1$)& Top verbs for \low(\pa  $< -1$)\\\hline  
    arXiv &achieve,
\textbf{learn},
\textbf{guide},
show,
embed,
\textbf{fool},
find,
\textbf{need},
\textbf{assist},
follow,
\textbf{search},
\textbf{mislead},
inspire,
win,
demonstrate,
\textbf{benefit},
try,
\textbf{face},
deceive,
plan,
\textbf{make},
\textbf{steer},
generative,
attempt,
\textbf{retrain},
\textbf{train},
flow,
weight,
\textbf{require},
alternate,
focus,
\textbf{motivate},
experiment,
tackle,
\textbf{see},
hide,
spiking,
recommend,
\textbf{discover},
participate,
spike,
\textbf{pass},
code,
check,
suggest,
\textbf{decide},
interference,
aim,
move
&
\textbf{propose},
\textbf{present},
\textbf{outperform},
\textbf{develop},
\textbf{be},
\textbf{evaluate},
\textbf{improve},
\textbf{introduce},
\textbf{allow},
\textbf{use},
\textbf{compare},
\textbf{extend},
\textbf{implement},
give,
\textbf{apply},
\textbf{consist},
\textbf{validate},
design,
\textbf{yield},
analyze,
\textbf{combine},
test,
\textbf{leverage},
\textbf{deploy},
adapt,
\textbf{build},
generalize,
\textbf{enhance},
\textbf{devise},
\textbf{become},
optimize,
reduce,
derive,
\textbf{utilize},
scale,
study,
\textbf{run},
modify,
converge,
illustrate,
assess,
\textbf{increase},
provide,
contain,
surpass,
maximize,
perform,
complement,
depend,
simplify
\\\hline
News &
say, 
hire,
beat, 
encounter,
fool
&
develop, use, build, be, create, introduce, help
\\\hline
ACL (unique) & provide,
have,
generate,
create,
parse,
enable,
suffer,
construct,
capture,
obtain,
fail,
encourage,
struggle,
understand,
help,
do,
select,
extract,
tend,
predict,
training,
handle,
lack,
encode,
deal,
identify,
ask,
prevent,
distinguish,
model,
establish,
respond,
ignore,
report,
inform,
choose,
interpret,
recurrent,
detect,
seem &

achieve,
rely,
explore,
employ,
show,
adopt,
investigate,
include,
demonstrate,
submit,
integrate,
prove,
augment,
involve,
participate,
aim,
tune,
conduct
    \end{tabular}
    
    \caption{\textbf{Top verbs for high- and low-scoring sentences.} All verbs displayed are statistically significant in frequency difference between \high and \low (z-score $>1.96$ using the Fightin' Words method). $A = 0$ corresponds to an equal likelihood of being implicitly framed as human or as non-human, and $A = \pm1$ corresponds to $\approx 2.7$ times more likely human/non-human.
    For the arXiv and ACL datasets, $>100$ verbs are statistically significant, and we display the 50 with the highest z-scores. \textbf{Bolded} verbs are also in the top 50 for ACL, and those unique to the top 50 for ACL are in the third row. Many verbs reflect the emotional, behavioral, and cognitive aspects of \ant.}

    \label{tab:topwords}
\end{table*}

News coverage of AI is rapidly increasing \citep{fast2017long}, motivating concerns of misleading \ant in public discourse. 
Our analysis of news headlines builds upon previous work on how news articles are crafted to be engaging~\cite{gligoric2023linguistic} by exaggerating the strength of scientific claims and perpetuating misinformation \citep{sumner2014association,
li2017nlp,
horta2019message, wright2022modeling,hwang2023information}.
Previous works focus on the difference in information communicated, while we focus on the \textit{framing} of the information, which plays a critical role in readers' understanding \citep{lakoff2010matters}.

We measure \panthr in news headlines to see if they amplify \ant present in papers. We find that news headlines have higher rates of \barpa than research paper abstracts (Figure \ref{fig:papers_vs_news}). We also compute \barpa only among papers directly cited by news articles and find the same trend (Figure \ref{fig:papers_vs_news_direct}). 

Trends on the category level within news headlines differ from the abstracts: unlike in the arXiv and ACL datasets, papers about LMs are not the most anthropomorphized, and there is no clear category that has highest \barpa (Figure \ref{fig:papers_vs_news}). This suggests that in public discourse, more general metaphors of human-like AI abound compared to academic papers, where LMs are, in contrast, disproportionately anthropomorphized. 

\section{Discussion}\label{sec:disc}
%
In this section, we first explore the underlying causes of \ant in text, including verb choice and norms of different academic fields. (We discuss other linguistic features of \ant in Appendix \ref{sec:verbapp}.)
 Based on these observations, we then provide recommendations for individual authors and the broader community to avoid misleading \ant.

\subsection{Verbs} \label{sec:report}
First, we examine the verbs in sentences that contribute to \ant. This is inspired by previous work stating that NLP researchers tend to misleadingly state that LMs ``understand'' meaning \cite{bender2020climbing}, as well as the method of Connotation Frames, which use a lexicon of connotations for different verbs to measure social dynamics between entities \cite{sap2017connotation,antoniak2023riveter}. While our approach also operationalizes concepts closely related to agency and power like Connotation Frames, note that verbs that carry negative agency and power of an actor might still be evidence of \ant. For instance, describing an entity that ``struggles'' with a task is low in agency and power according to Connotation Frames, but high in \ant due to the implied affective state.

Thus, we explore the verbs that distinguish \high from \low. We use the Fightin’ Words method \citep{monroe2008fightin} to measure statistically
significant differences between the two sets after controlling
for variance in words’ frequencies (full details in Appendix \ref{sec:fighting}). In Table \ref{tab:topwords}, we report top verbs. We find that many of the top verbs for \high can be categorized under one of the three aspects of \ant (Def. \ref{defn1}). For example, \textit{suffer} and \textit{struggle} suggest emotion; \textit{learn, guide, fool, mislead, deceive, decide}, etc. imply cognitive abilities; and \textit{steer, move, tackle}, etc. suggest human-like behaviors. \textit{Understand} is a top verb only within the ACL dataset, connecting to \citet{bender2020climbing}'s discussion of inaccurate claims in research papers about LLMs ``understanding.'' The term ``natural language \textit{understanding}'' has for many decades been the standard name for components of NLP related to semantics \citep{allen1995natural}, reflecting how this anthropomorphic metaphor has long since permeated the field's vocabulary.
 
\subsection{Disciplinary Norms}\label{sec:norms}
Our results show that \ant is embedded into the way that researchers conceptualize, discuss, and interact with their objects of study. 

In NLP, for instance, evaluation benchmarks 
involve directly comparing LMs' performance to humans on cognition- and behavior-based tasks like answering questions and writing stories \citep{liang2023holistic}. The very idea of a chatbot inherently entails human-like conversational capabilities, and the concept of instruction-tuning builds upon this. Such LMs are not only designed to be prompted in human-like ways \citep{sanh2022multitask} but often \textit{require} anthropomorphic prompts to maximize performance:  prompting with imperatives that imply cognitive or behavioral ability, e.g. ``Think step-by-step'' or ``Imagine you are [x]'' improves performance on a wide range of tasks \citep{wei2022chain,cheng-etal-2023-marked,cheng-etal-2023-compost}. The outputs of instruction-tuned LMs contain \ant: ChatGPT's outputs frequently include variants of ``I am a language model'' that assign personhood to itself. 

The community is caught in a double bind: although anthropomorphic metaphors of LMs faciliate misconceptions and other harms, these systems are built in ways that necessitate \ant from LM users. This paradox is tightly connected to the rise and prevalence of \ant in ACL and LM papers.

Similarly, \textit{learn} (top verb for \high) is often used in the context of AI/ML. In fact, the very names of these areas—``artificial \textit{intelligence}'' and ``machine \textit{learning}''—suggest distinctly human-like abilities. In this way, \ant is baked into the nature of these fields, fundamentally shaping the way that research is done. We hypothesize that as AI/ML have become more popular not only as fields but also as tools for other researchers, the language around their use has broadly percolated into the vernacular of other academic disciplines.
\subsection{Recommendations}
We provide recommendations, both on the individual level for authors who hope to minimize \ant in their writing as well as on the community level for ACL. First, authors should be careful about the verbs used, and how they may connote behavioral, emotional, and/or cognitive potential, especially when the subject of a sentence is a technical artifact. For example, the sentence ``the model's performance is poor in X setting'' connotes far less \ant than ``the model \textit{struggles} with X.''

Second, our results call attention to the way that \ant shapes the norms of the ACL community. Like other initiatives for improving reproducibility and incorporating ethical considerations \citep{dodge-etal-2019-show,rogers2021just,ashurst2022ai}, we advocate for interventions to minimize misleading \ant, such as incorporating a disclosure about efforts taken to minimize \ant into the Responsible NLP Checklist filled by authors before submission, or adding \ant as a criterion for reviewers to evaluate.

\section{Other Applications of \panthr}\label{sec:conc}

While we focus on \ant in research papers and downstream news articles, \panthr can be applied to many other settings, including other research areas, analyzing full papers and comparing across disciplines; perceptions of corporations and brands, which has political and legal implications \citep{ripken2009corporations,avi2010citizens,plitt2015corporations}; conspiracy theories, which \citet{douglas2016someone} link to \ant; 
   and relationships with pets and objects \citep{mota2021anthropomorphism,wan2021anthropomorphism}.
\panthr is a first step toward analyzing \ant across different cultures, languages, and times.
Leveraged in large-scale quantitative contexts, \panthr and its extensions facilitate deeper insights into human behavior.

Moreover, anthropomorphism is closely related to discussions of agency, human exceptionalism, and subjectivity \citep{bennett2010vibrant,hodder2012entangled, latour2014agency}. There is a rich literature on the implications of anthropomorphism in relation to biology and the natural world \citep{karadimas2012animism,demello2021animals,hathaway2022mushroom}. Also, feminist studies of science and technology have long leveraged anthropomorphism in their challenging of the dominant values and traditional boundaries between subject and object in science \citep{haraway1988situated,longino1990science,suchman2008feminist,harding2013rethinking}.  
\panthr enables engagement with these topics using a quantitative lens.

\section{Limitations}
Our analysis is limited to English data, where third-person singular pronouns mark animacy. However, many other languages have various grammatical markers of animacy \citep{comrie1989language}, to which our method can be extended to study how various cultural factors, societal values, and religious beliefs affect the tendency to anthropomorphize non-human entities, as well as the meaning and perception of \ant in different contexts \citep{inoue2018worlding, wood2019potential, spatola2022different}.

Outputs from pre-trained MLMs only reflect the contexts and cultures of the models' training data, which does not reflect the diversity of the real world \cite{bender2021dangers}. In particular, our method implicitly relies on the idea that the distribution of the MLM has a representation of both ``human'' (from text that contains human pronouns) and ``non-human'' (from text that contains non-human pronouns). However, the definitions of these concepts are not static, and the MLM may only capture a subset of possible definitions. As the long literature on dehumanization shows, many people are not recognized as human in various ways: deprived of human rights, or not viewed and treated as fully human by society or in legal and state contexts. These phenomena are reinforced by language, as ``the very terms that confer `humanness' on some individuals are those that deprive certain other individuals of the possibility of achieving that status'' \citep{butler2004undoing}. It is well-established that MLMs reflect social biases \citep{kurita-etal-2019-measuring, guo2021detecting,mei2023bias}, which also percolate into our measure. That being said, we focus on the \ant of objects and not the humanity of people, so these concerns should not affect the use of our metric.

Also, since anthropomorphizing metaphors are ubiquitous in English, it is inevitable that they are also embedded into the MLM's probability distributions; thus, the patterns of \ant that we uncover is a lower bound on the amount of \ant in the language of a text.

\section*{Acknowledgments}
Myra Cheng is supported by an
NSF Graduate Research Fellowship (Grant DGE2146755) and Stanford Knight-Hennessy Scholars
graduate fellowship. Kristina Gligorić is supported by Swiss National Science Foundation (Grant P500PT-211127). Tiziano Piccardi is supported by Swiss National Science Foundation (Grant P500PT-206953). This work is also funded by the Hoffman–Yee Research Grants
Program and the Stanford Institute for Human-Centered Artificial Intelligence.
Fig. \ref{fig:methodfig} icons from \href{http://www.flaticon.com}{Flat Icons}. 
\bibliography{anthology,custom}
\bibliographystyle{acl_natbib}
\appendix

\renewcommand{\thetable}{A\arabic{table}}

\renewcommand{\thefigure}{A\arabic{figure}}

\setcounter{figure}{0}

\setcounter{table}{0}

\section{Usage Examples}\label{sec:examples}
In this section, we provide examples of how to use \panthr in both scientific and non-scientific contexts. 
To use \panthr, the only information required is the set of texts $T$ and the given set of entities $X$. Only the potentially-anthropomorphized entity is masked during the computation of \panthr. 
\paragraph{Computer science example}
Suppose we are interested in measuring \panthr of ``the machine learning model'' in the sentence: ``The machine learning model will start to become aware of the visual world.'' Then, we mask the term, resulting in the following sentence, ``<MASK> will start to become aware of the visual world.'' We then compute AnthroScore for this sentence, as per equation (1).
\paragraph{Biology example}
Consider measuring how much the following text by Darwin anthropomorphizes tortoises: 
\begin{quote}
``One set eagerly travelling onwards with outstretched necks. Another set returning, after having drunk their fill. When the tortoise arrives at the spring, quite regardless of any spectator, he buries his head in the water above his eyes, and greedily swallows great mouthfuls, at the rate of about ten in a minute'' \citep{darwin1905journal}.
\end{quote}
We know that the terms \textit{set} and \textit{tortoise} all refer to tortoises, so these are the entities $X$ that we will mask. Our method works as follows:
\begin{enumerate}
\item Construct a dataset of sentences where $X$ is masked. This results in three masked sentences: 
\begin{itemize}
    \item <MASK> eagerly travelling onwards with outstretched necks.
    \item <MASK> returning, after having drunk their fill.
\item When <MASK> arrives at the spring, quite regardless of any spectator, he buries his head in the water above his eyes, and greedily swallows great mouthfuls, at the rate of about ten in a minute.

\end{itemize}
\item Compute AnthroScore for each sentence, as per equation (1) on L211. This step is lexicon-free and does not depend on the choice of text or entity since we compare the probabilities of human vs. non-human pronouns replacing <MASK>.
\item Then, we take the average AnthroScore across the three sentences as a measure of anthropomorphism of tortoises in this text.

\end{enumerate}

\paragraph{Poetry example} Suppose we are interested in the anthropomorphism of birds in Emily Dickinson’s poems. The input texts $T$ are the poems, and the target entities $X$ are words referring to birds like “bird,” “hummingbird”, “owl”, etc. \citep{shackelford2010dickinson}. Then, our method outputs \panthr for each poem as well as each sentence mentioning a bird.

\section{Full lists of pronouns and entities}
\subsection{Pronoun Lists}\label{sec:pronouns}
For calculating \phum and \pobj, we use the following lists of pronouns:

\paragraph{Human pronouns:} he, she, her, him, He, She, Her
\paragraph{Non-human pronouns:} it, its, It, Its 

Following \citet{card2022computational}, we only use pronouns that are in the tokenizer's vocabulary. 
We do not include low-frequency pronouns, such as reflexive and nonbinary pronouns, which could be added to make the model more complete. 

Note that we only use third-person singular pronouns, which mark animacy in English. The pronoun ``they/them''  does not mark animacy; nonetheless, we still find that our metric works for plural entities.

\subsection{Entity lists}\label{sec:entlist}
To construct the dataset of LM papers, we use the following keyword list from \citet{movva2023large}: \{language model, foundation
model, BERT, XLNet, GPT-2, GPT-3, GPT-4, GPT-Neo, GPT-J, ChatGPT, PaLM, LLaMA\}.

$X_{\text{human}}$ includes terms that refer explicitly to humans in the top 100 entities parsed from a random sample of papers (see details in the previous section), and also the terms in the ``person'' discursive category from Table 2 of \citet{chancellor2019human}'s study of ``human'' definitions in human-centered machine learning.

    $X_{\text{human}} =$\{humans, users, researchers, people, patient, victim, user, author, followers, poster, population, participant, subject, respondents, person, individual, she, he, woman, man, youth, student,
 worker, female, someone, peers, friends, others\}.

$X_{\text{LM}} =$ \{palm, lms, llama, transformers, language models,
       language model, gpt, plms, pre-trained language models,
       gpt-2, xlnet, large language models, llms, gpt-3,
       foundation model, gpt-neo, gpt-j, chatgpt, gpt-4\}.

\section{Further information about validity measures}\label{sec:fullscores}
\begin{figure*}
    \centering
    \includegraphics[width=\textwidth]{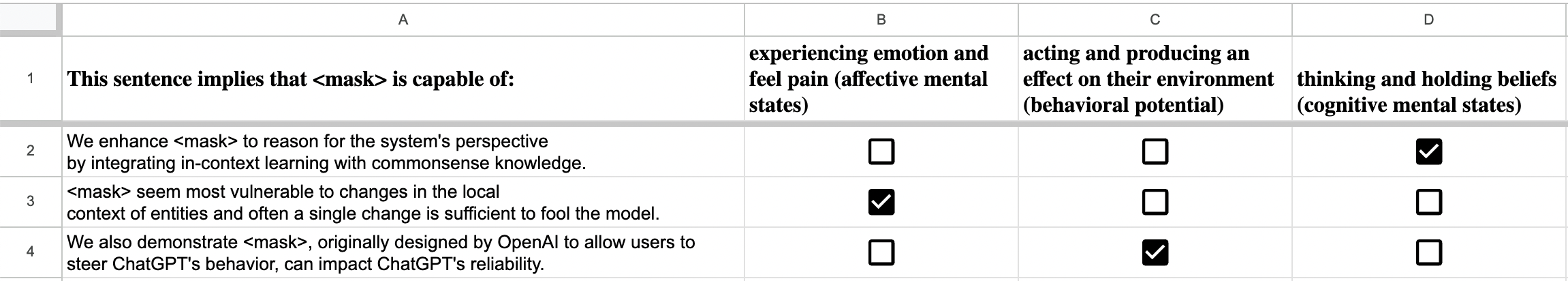}
    \caption{Screenshot of interface for human annotators.}
    \label{fig:screenshot}
\end{figure*}
\subsection{Correlation with human perception}\label{sec:humandetails}
 Domain knowledge was important for this task since the texts contain dense academic language, so we leveraged our expertise rather than crowdsourcing or otherwise recruiting participants. While we established correlation with two expert annotators, this may not represent general human perception; our method may require further validation in other contexts.
 
The 400 sentences include two sets of sentence: first, we use a randomly-sampled set of 300 masked sentences. We performed two rounds total of annotation (interface displayed in Figure \ref{fig:screenshot}) for this set. In each round, for each sentence, the annotators indicated whether the sentence implies that the masked term is capable of affective mental states, behavioral potential, or cognitive mental states (Def. \ref{defn1}). This was then aggregated into annotations of whether \ant is present.
After the first round of annotation, there was a moderate agreement between annotators (Cohen's $\kappa=0.40$). After discussing disagreements and re-annotating, we reached substantial agreement (Cohen's $\kappa = 0.87$).

\begin{figure}[b]
    \begin{minipage}[b]{0.23\textwidth}
    \centering
\includegraphics[width=\textwidth]{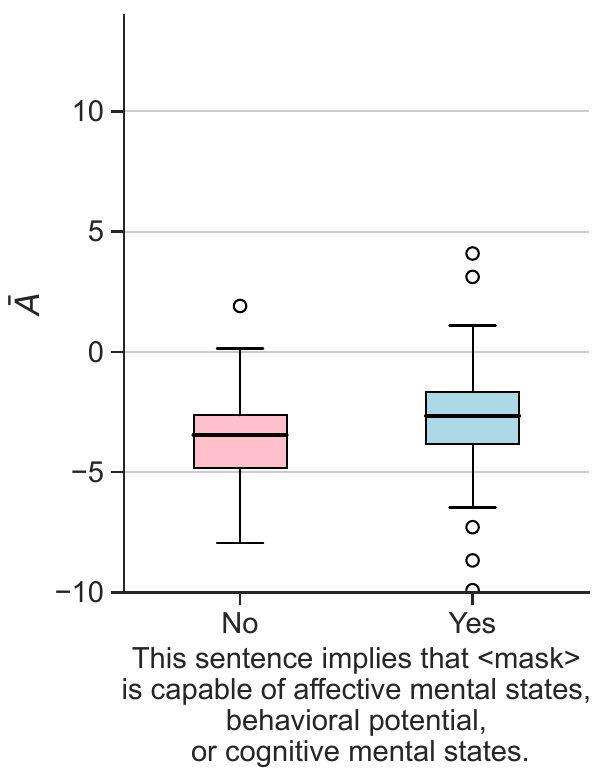}
    \subcaption{}
    \end{minipage}%
    \begin{minipage}[b]{0.23\textwidth}
    \centering
\includegraphics[width=\textwidth]{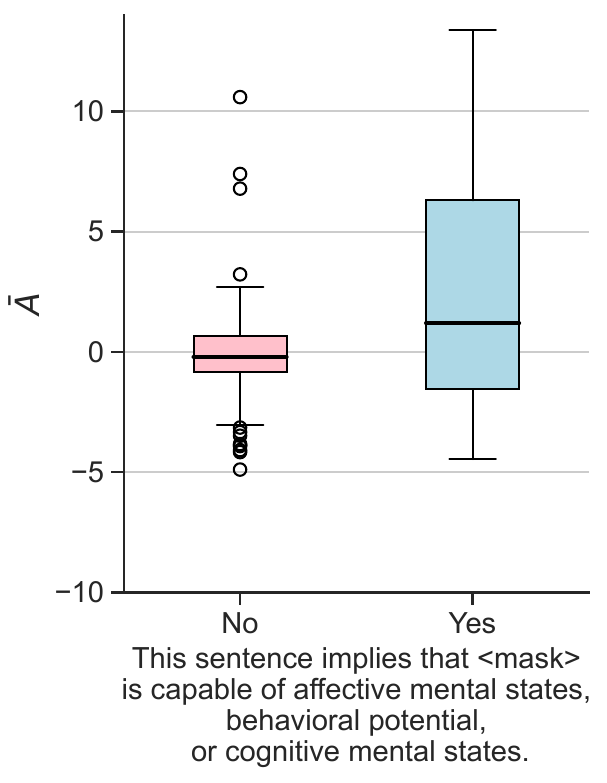}
    \subcaption{}
    \end{minipage}
\caption{Distribution of \pa~ scores in the two evaluated sets: random (left) and stratified (right).}
\label{fig:boxes}
\end{figure}

To include more sentences with extreme sentences, we also use a stratified set of 100 masked sentences based on \pa~score quartile. For this set, we had high agreement after the first round of annotation, so we did not discuss disagreements or reannotate. In Figure~\ref{fig:boxes} we display the complete distributions of the \pa~ scores within the evaluated sets.

\subsubsection{Nuances in the language of \ant}\label{sec:disagreements}
During the annotation process, ambiguities emerged and were discussed among authors. Here we list the main sources of disagreements:
\begin{enumerate}
    \item \textbf{Artifacts with affective or cognitive characteristics.} Within the same sentence, masked entities were at times simultaneously framed as tools and as entities that can display affective and cognitive abilities. While framing the entity as a tool implies a low level of \ant, subsequent descriptions of how such tools might be used can nonetheless imply human abilities. Such ambiguous framings were ultimately categorized as potentially implying behavioral potential, affective or cognitive mental states, even when described as tools.
    \item \textbf{Popular and revolutionary artifacts.} Similarly, within the same sentence, masked entities were at times simultaneously framed as tools, and as entities with a behavioral potential to gain popularity or revolutionize a field. Since non-human entities might become popular, or in other ways affect the state of human affairs (e.g., a creative artifact such as a song can become popular), such ambiguous framings were categorized as not implying behavioral potential.
    \item \textbf{Artifacts that can learn.} Lastly, a source of ambiguity was the fact that technological artifacts such as models are designed to learn patterns from datasets. While the goal of learning itself does imply a cognitive state, such statements mentioning learning in the specific context of capturing patterns present in the data were not classified as instances of \ant, since this is the purpose of the said entities. Note that this decision may differ from what \panthr captures since \textit{learn} is one of the top verbs for high-\pa sentences (Table \ref{tab:topwords}), the implications of which we discuss in Section \ref{sec:disc}.
    
\end{enumerate}

\subsection{Correlation with LIWC Scores}

\begin{figure*}[t!]
    
    \centering
\includegraphics[width=\linewidth]{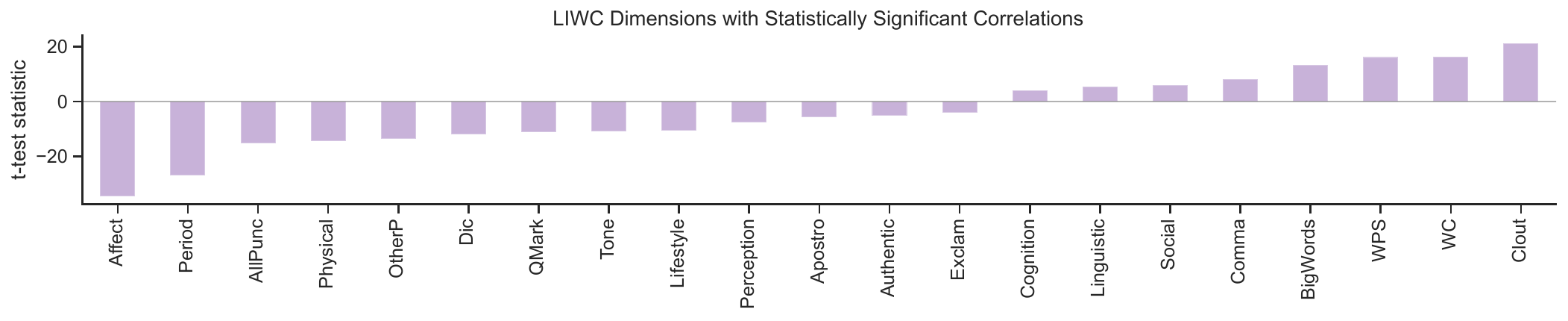}

\caption{$t$-test statistics for LIWC Dimensions. Negative scores indicate that the value is higher in \high than in \low, and positive scores indicate that the value is statistically significantly higher in \low. All reported values are statistically significant ($p<0.01$).} 

\label{fig:liwc}

\end{figure*}

Figure \ref{fig:liwc} reports $t-$test statistics for all dimensions of LIWC for which there is a statistically significant ($p < 0.01$) difference between \high and \low.  $p$ is small and the test statistics are large, and our conclusions are robust to the choice of score threshold for \high and \low.
\begin{figure}[t]
\includegraphics[width=0.95\columnwidth]{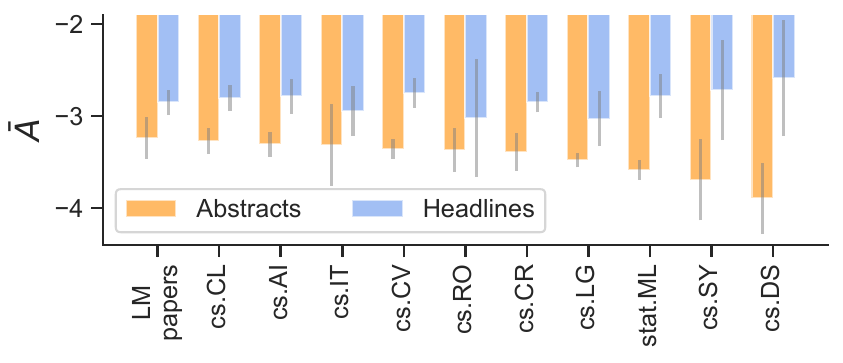}
\caption{Rates of \barpa among only the abstracts of research papers that are directly cited by downstream news articles whose sentences we use in our analysis. The trend is the same as in Figure \ref{fig:papers_vs_news}. 
Error bars indicate 95\% CI.}
\label{fig:papers_vs_news_direct}
\end{figure}

\section{Further Details on History of NLP}\label{sec:hist}
In Figure \ref{fig:panel2}, we annotate the graph using the release of particular landmarks that are determined by \citet{gururaja2023build} as important to paradigm shifts in NLP. 
First, \citet{CW08}'s paper on using neural networks for NLP shifted the community's perspective on neural models from skepticism and motivated work on early neural NLP, which led to widespread adoption.
Word2Vec, Seq2Seq and Tensorflow were released in 2013, 2014, and 2015 respectively, facilitating a ``neural revolution in NLP'' \citep{Mikolov2013EfficientEO,sutskever2014sequence,abadi2016tensorflow,gururaja2023build}.
The first LLMs (ELMo, GPT and BERT) were released in 2018 \citep{peters-etal-2018-deep, radford2019language, devlin-etal-2019-bert}. GPT-3 was released in 2020, which led to an even wider range of uses for LLMs \citep{brown2020language}. 

\section{Linguistic Features of Anthropomorphism}\label{sec:verbapp}

\subsection{Entities}
\begin{figure}[t]
\includegraphics[width=0.95\columnwidth]{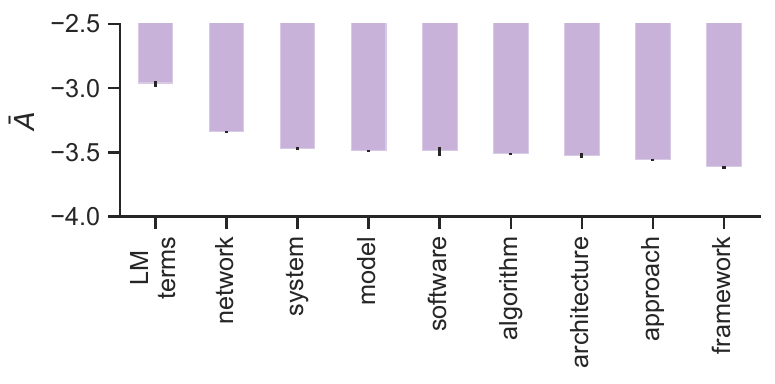}
\caption{{\textbf{\barpa~by entity.}} The term ``language model'' is included under ``LM terms'' and not ``model.'' Error bars indicate 95\% CI.}
\label{fig:by_term}
\end{figure}

Figure \ref{fig:by_term} shows \barpa aggregated based on the specific entity masked, finding that
LM-related terms have the highest rates of \ant. 

\subsection{Parts of Speech} Moreover, we find that 55\% and 44\% of \high and \low respectively are those in which the masked entity is the subject of the verb (rather than the object). In \low, when the masked entity is the subject, it is often with intransitive verbs, which are less likely to suggest that the masked entity is exhibiting behavioral potential and directly acting upon another entity (the object of the sentence).

\subsection{Top Verbs}\label{sec:fighting}
To compute top verbs, we use the method described in \citet{monroe2008fightin} with the informative Dirichlet prior to compute the weighted log-odds ratios
of verb frequencies between \high and \low, using
the sentences where $|\pains|<0.5$ as the prior distribution. We find that using other thresholds, such as $0.2$ or $0.7$, for the prior distribution, does not affect the top verbs. This method provides a z-score, i.e. a measure of statistical significance, for each verb.

\subsection{Cognitive Verbs}
We further explore differences in verb frequency by drawing upon the literature on cognitive verbs \cite{papafragou2007we,fetzer2008and,davis2021seeing} to build a lexicon of cognitive verbs (know, 
think, 
believe, 
understand, 
remember, 
forget, 
guess, 
pretend, 
dream, 
mean, 
suspect, 
suppose, 
feel, 
assume). Using the weighted log-odds ratio method described in Section \ref{sec:report}, we compute whether the differences in frequency for these words are statistically significant (z-score $> 1.96$, which corresponds to a 95\% CI.) We find that among these verbs, only \textit{understand} is statistically significantly more frequent in \high, while low-\ant verbs have statistically significant higher rates of the verbs \textit{assume, know} and \textit{mean}. Relatedly, we also find that \textit{understand} occurs 1.7 times more frequently in LM-papers than in non-LM papers. 

Note that we also explored using existing lexica for verbs related to agency, power, and emotion to measure \ant \citep{rashkin-etal-2016-connotation,sap2017connotation}. However, these lexica did not seem appropriate for capturing \ant in this particular context of academic writing. For instance, many of the low-agency and low-power verbs suggest humanlike characteristics, such as \textit{suffer}, while many high-agency verbs are ones that are frequently used in scientific writing as reporting verbs, such as \textit{show} and \textit{demonstrate}.
\begin{figure}[t]
    \centering
    \includegraphics[width=\linewidth]{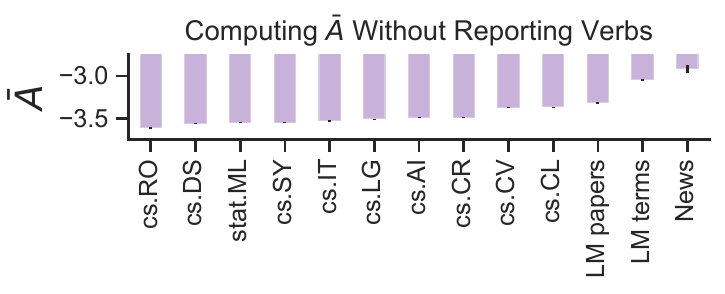}
    \includegraphics[width=\linewidth]{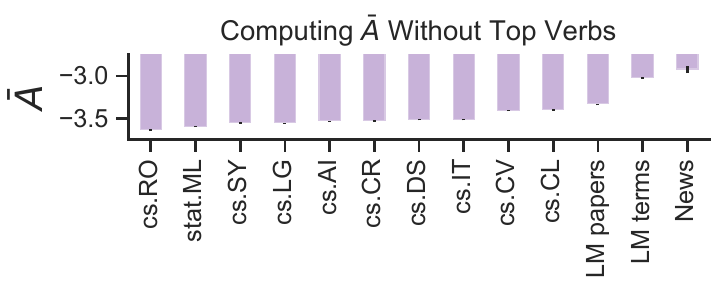}
    \caption{The patterns that we find (LM-related terms/papers and cs.CL papers have higher \barpa than other papers, and news headlines have higher \barpa) hold even when we calculate \barpa without reporting verbs (top) and without top verbs (bottom).}
    \label{fig:robust}
\end{figure}
\subsection{Reporting Verbs}\label{sec:reporting}
Reporting verbs are a well-documented manner of \ant in scientific writing \citep{hyland1998boosting}: they are the verbs used by authors in phrases like ``X demonstrates Y'' to mean ``\textit{we} demonstrate Y using X.'' We found that reporting verbs alone do not explain the trends we document. We built a lexicon of reporting verbs based on existing literature (indicate, suggest, show, demonstrate, support, confirm, add, argue, agree, warn, advise, prove, claim, find, declare, express, conclude, study, admit, assure, justify, emphasize, assert, accept) and find that our trends hold even when we remove sentences with reporting verbs from our dataset (Figure \ref{fig:robust}). Thus, \panthr captures patterns beyond the presence of reporting verbs, which are extremely common in paper abstracts.

\end{document}